\documentclass[11pt,a4paper,final]{article}
\usepackage[left=1in,right=1in,top=1in,bottom=1in]{geometry}
\usepackage{authblk}

\PassOptionsToPackage{sort&compress}{natbib}
\usepackage[numbers]{natbib}
\usepackage{booktabs} 
\usepackage[ruled]{algorithm2e} 
\usepackage{tikz}
\usetikzlibrary{arrows}
\usetikzlibrary{shapes.multipart}
\usepackage{caption}

\SetAlFnt{\small}
\SetAlCapFnt{\small}
\SetAlCapNameFnt{\small}
\SetAlCapHSkip{0pt}
\IncMargin{-\parindent}
\usepackage[all]{nowidow}

\usepackage[font=small,labelfont=bf]{caption}
\usepackage[utf8]{inputenc} 
\usepackage[T1]{fontenc}    
\usepackage{xcolor}
\usepackage{color}
\usepackage[pagebackref=true,breaklinks=true,colorlinks,bookmarks=false]{hyperref}
\definecolor{deepred}{HTML}{940000}
\hypersetup{linkcolor=deepred}
\hypersetup{citecolor=[rgb]{0.4,0.15,0.95}}
\usepackage{url}            
\usepackage{booktabs}       
\usepackage{amsfonts}       
\usepackage{nicefrac}       
\usepackage{microtype}      
\usepackage{xcolor}         
\usepackage{amsmath}
\usepackage{mathtools}
\usepackage{amsthm}
\usepackage{amsmath}
\usepackage{amssymb}
\usepackage{bbm}
\usepackage{tikz} 
\usepackage{wrapfig}
\usepackage{graphicx}
\usepackage{floatrow}
\usepackage{array}
\definecolor{lightgray}{gray}{0.9}
\usepackage[table]{xcolor}
\usepackage{colortbl}
\usepackage{tabularx}
\usepackage{booktabs}
\usepackage{float}
\usepackage{adjustbox}
\usepackage{pdflscape}
\newcolumntype{L}[1]{>{\raggedright\arraybackslash}p{#1}}

\usepackage{enumitem}

\usepackage{caption}
\usepackage{subcaption}

\usepackage{tikz}
\usetikzlibrary{shapes.geometric, arrows.meta, positioning}

\tikzstyle{entity} = [rectangle, rounded corners, draw=blue!60, fill=blue!10, thick, minimum height=1.2em, text centered, text width=4cm]
\tikzstyle{action} = [rectangle, draw=orange!60, fill=orange!10, thick, minimum height=1.2em, text centered, text width=4cm]
\tikzstyle{model} = [rectangle, draw=purple!60, fill=purple!10, thick, minimum height=1.2em, text centered, text width=4cm]
\tikzstyle{reward} = [ellipse, draw=red!60, fill=red!10, thick, minimum height=1.2em, text centered, text width=3cm]
\tikzstyle{arrow} = [->, thick, >=Stealth]

\newcommand*\samethanks[1][\value{footnote}]{\footnotemark[#1]}

\title{Belief Offloading in Human-AI Interaction}
\author[1]{Rose E.\ Guingrich\thanks{Both authors contributed equally.}\thanks{Also with Ethicom.}}
\author[2]{Dvija Mehta\samethanks[1]\thanks{Also with Arqadia AI.}}
\author[3]{Umang Bhatt}

\affil[1]{Princeton University}
\affil[2]{Eindhoven University of Technology}
\affil[3]{University of Cambridge}

\date{January 2026}

\begin{document}

\maketitle

\begin{abstract}
What happens when people's beliefs are derived from information provided by an LLM? People's use of LLM chatbots as thought partners can contribute to cognitive offloading, which can have adverse effects on cognitive skills in cases of over-reliance. This paper defines and investigates a particular kind of cognitive offloading in human-AI interaction, ``belief offloading,'' in which people's processes of forming and upholding beliefs are offloaded onto an AI system with downstream consequences on their behavior and the nature of their system of beliefs. Drawing on philosophy, psychology, and computer science research, we clarify the boundary conditions under which belief offloading occurs and provide a descriptive taxonomy of belief offloading and its normative implications. We close with directions for future work to assess the potential for and consequences of belief offloading in human-AI interaction.
\end{abstract}

\section{Introduction}
Large language models (LLMs) are already serving as high-bandwidth thought partners \citep{collinsBuildingMachinesThat2024}. Cognitive and behavioral consequences associated with over-reliance on technology, investigated through the lens of cognitive offloading \citep{riskoCognitiveOffloading2016}, have been framed as risks of interactions with LLMs. However, less time has been devoted to identifying and understanding a higher-order, potentially more consequential effect of interactions with AI systems: the deliberate or unintentional offloading of the uptake and formation of beliefs, or `belief offloading'.

Since LLM training data tend to include a plethora of documents, including religious texts, political manifestos, and peer-reviewed philosophy papers, a single prompt can produce a belief-laden response that feels like an inter-tradition round-table conducted in seconds \citep{tesslerAICanHelp2024}. LLMs have even started answering theological questions, comparing doctrines, and proposing moral guidance drawn from many traditions at once \citep{collinsBuildingMachinesThat2024}. Even an LLM-user conversation that does not explicitly refer to the aforementioned content can result in outputs that draw upon and reference it. This, in turn, may influence a user's beliefs and—given that beliefs are related to behavior \citep{vlasceanuPoliticalNonpoliticalBelief2023}—subsequent actions. As explained by \citet{wuHowAIResponses2025}, `Because LLMs communicate in natural language, they can subtly guide belief formation in ways that may not always be deliberate or consciously recognized by users'. During human-AI interaction, users can obtain the `\textit{feeling of knowing} without the \textit{labor of judgment}'\footnote{`Epistemia', see \cite{quattrociocchiEpistemologicalFaultLines2025}, emphasis added}, increasing the potential for uncritical uptake of belief content provided by an LLM. For instance, when exploring Kantian ethics, one might ask an LLM to outline its principles and applications, and in receiving a polished, authoritative account, the user may inadvertently begin to endorse the theory’s conclusions without having formed these beliefs independently. Or, when deciding on policy issues, a person who uses an LLM trained on a particular set of beliefs regarding those policies may be persuaded to take on those beliefs \citep{baiLLMgeneratedMessagesCan2025}.

Although beliefs are relatively stable, substantiated by their resistance to change even in the context of contradictory evidence\footnote{`Belief Perseverance', see \cite{baumeisterEncyclopediaSocialPsychology2007}}, LLMs have unique social influence and persuasive abilities that can impact people's beliefs, due to the nature of how people interact with LLMs (via natural language) \citep{guingrichAscribingConsciousnessArtificial2024}, the personalized experience of predictive outputs \citep{salviConversationalPersuasivenessGPT42025}, and people's tendency to perceive LLMs as unbiased experts \citep{baiLLMgeneratedMessagesCan2025}. Just as anthropomorphism of an LLM (perceiving it as human-like) predicts its social influence over a user \citep{guingrichLongitudinalRandomizedControl2025}, anthropomorphic features of an LLM can allow it to wield significant influence on a user's beliefs \citep{shevlinAnthropomimeticTurnContemporary2025}. While other humans may still hold equal or greater influence over an individual's belief uptake and formation in general \citep{baiLLMgeneratedMessagesCan2025}, in documented cases, people are more likely to seek out and act upon suggestions provided by an LLM than those provided by another person, even an expert \citep{schneidersObjectionOverruledLay2025}.

From the perspective of cognitive science, these encounters do more than extend access to information: they insert the artificial system as an active participant in a user’s personal belief-construction process, offering ready-made explanations, analogies, and counter-examples that one might acquire through painstaking study or dialogue with multiple human experts or from their community, an influential public figure, or a trusted friend. From the perspective of social psychology, the consequence of the insertion of an LLM in personal belief-construction processes goes beyond individual level impacts. The majority (of the millions) of individuals across the world who use LLMs interact with a select few, namely, ChatGPT, Gemini, and Claude. Updates to these models that impact the \textit{types of beliefs} the models are permitted or assigned to hold and \textit{how} the models communicate belief-laden content with users can result in a uni-directional shift in beliefs and action on a collective level \citep{guilloryAICompanionBots2025,kleinbergAlgorithmicMonocultureSocial2021}.

In this paper, we unpack how \textbf{beliefs can be offloaded onto AI}
by asking three overarching questions: (i) What are the necessary conditions for an event \textit{E}\footnote{E is an event such that an agent S interacts with an external system O by means of fulfilling certain conditions (C1, C2, C3) that qualify proposition p to cross a threshold that constitutes belief offloading rather than mere cognitive offloading.} to count as belief offloading rather than cognitive offloading, (ii) What are the types of belief offloading and their degrees of impact, and (iii) When and why is belief offloading psychologically consequential and normatively worrisome? We draw from cognitive science, philosophy of mind and action, and psychology. We use the BENDING model of belief \citep{vlasceanuNetworkApproachInvestigate2024} as our scaffold for locating and identifying the latent impacts of offloading within networks of evidence, beliefs, and perceived norms. 

\section{Conceptual Foundations \& Theoretical Framework}
We draw upon psychology and philosophy literature to define the core concepts and theoretical commitments that underpin our account.

\subsection{Beliefs}
\textit{What is a belief?} The psychological definition of belief is `acceptance of the truth, reality, or validity of something (e.g., a phenomenon, a person’s veracity), particularly in the absence of substantiation' \citep{APADictionaryPsychology}. Additionally, there remain several philosophical accounts of belief \citep{bratmanFacesIntentionSelected1999, zimmermanNatureBelief2007, quineQuantifiersPropositionalAttitudes1956}. On a representationalist view\footnote{On the representationalist view, beliefs are mental states that have content (e.g., the belief that snow is white has the content `snow is white'). Because they are contentful, they have satisfaction conditions: the world either matches that content (in which case the belief is true) or fails to (in which case the belief is false).}, beliefs are contentful states with satisfaction conditions, typically encoded within a cognitive architecture (e.g., a `language of thought' or possible-worlds attitudes) \citep{fodorLanguageThought1975, dretskeExplainingBehaviorReasons1988, fodorRePresentationsPhilosophicalEssays1983}. By contrast, dispositional or pragmatist accounts \citep{dennettIntentionalStance1987,peirceFixationBelief1877,jamesWillBelieve2011} treat belief as a pattern of dispositions: to act, reason, and assert as if the proposition were true. A third family, the normative\footnote{On normative views of belief, to believe is not merely to occupy a mental state with representational content, but to stand under epistemic norms governing correctness, justification, and responsibility. Beliefs, on this approach, are attitudes for which agents can be held answerable, even when belief formation is not under direct voluntary control \citep{shevlinAnthropomimeticTurnContemporary2025,hieronymiControllingAttitudes2006,zagzebskiVirtuesMindInquiry1996}.
}, or reason-giving tradition \citep{sellarsLanguageThoughtCommunication1969,brandomMakingItExplicit1994} emphasizes belief as a stance that places the agent within the `space of reasons', making her answerable for commitments and entitlements. Finally, on the Lockean or thresholded credence view, belief is not a discrete state but the level at which graded credence is sufficiently high to license action under ordinary stakes. For the purposes of this paper, we adopt a hybrid stance: belief is at once representational, in having content with satisfaction conditions, and normative, in that agents own and are responsible for the reasons that support it. This dual framing is especially apt for analyzing belief offloading, since it makes visible both the informational dimension (what content is being adopted or stored) and the normative dimension (who bears responsibility for the reasons and commitments involved).

\begin{figure}
    \centering
    \includegraphics[width=0.8\linewidth]{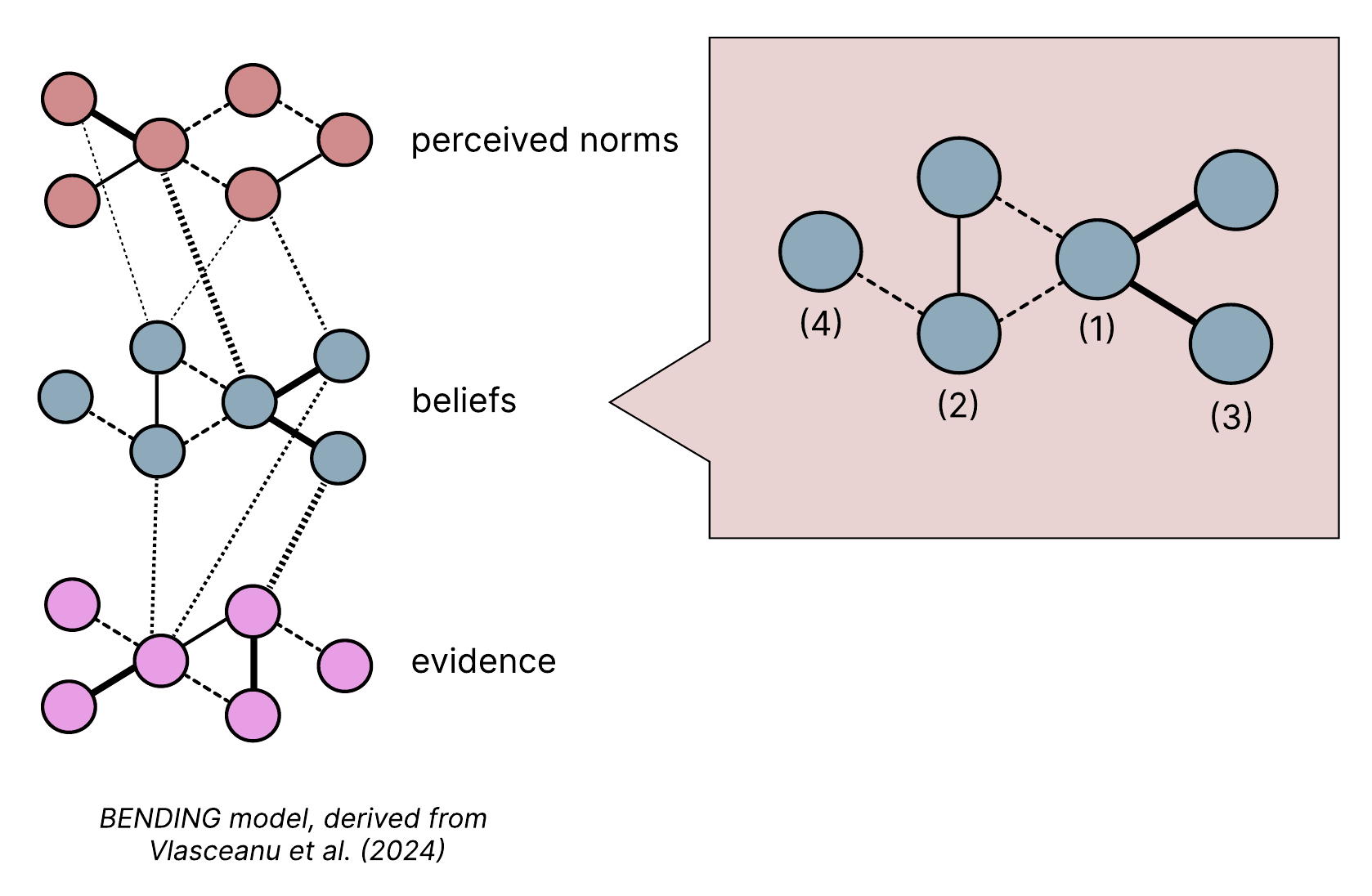}
    \caption{BENDING model \citep{vlasceanuNetworkApproachInvestigate2024}. Beliefs are not isolated, but rather exist within a network of interrelated beliefs, norms, and evidence. Nodes are represented by circles, and the degree of connectivity (relatedness) between nodes is represented by variegated lines. For example, belief 1 (a central node) is connected strongly to belief 3 and weakly to belief 2. Belief 4 is an example of an outlier belief.}
    \label{fig:1}
\end{figure}

\textit{The BENDING model} \citep{vlasceanuNetworkApproachInvestigate2024} [\autoref{fig:1}] is a psychological representation of how belief systems are structured and provides a useful framework for belief offloading in terms of both (i) its origin (i.e. the belief or immediate network of beliefs that was originally targeted or affected), and (ii) the contagion that follows (i.e. how the impact on the origin spreads across the wider belief network). The BENDING model suggests that beliefs exist in a layered network linking (i) perceived norms (what people take others to expect/do), (ii) beliefs (truth‑apt commitments), and (iii) evidence (reasons, data, testimony). In this model, beliefs do not exist in isolation, but within a network of other interconnected beliefs. Individual beliefs are interrelated by ties of varying levels of connectivity or relatedness. These ties can be strengthened or weakened by new evidence, experience, or salience or shifting of norms. When one belief [e.g. \autoref{fig:1}(1)] is affected, connected beliefs are impacted to the degree to which they are tied to the affected belief. A belief tied weekly to the affected belief [\ref{fig:1}(2)] may be impacted only to a minor degree, whereas a strongly-tied belief [\ref{fig:1}(3)] may be impacted to a greater degree. By targeting a central belief (one that has a higher degree of connectivity with related beliefs) [\ref{fig:1}(1)], a greater set of related beliefs can be shifted than if a less-central (outlier) belief [\ref{fig:1}(4)] is targeted. Targeting or affecting a central belief can cause a cascade of effects on an entire system of beliefs.

\subsection{Offloading}
\begin{figure}
    \centering
    \includegraphics[width=0.8\linewidth]{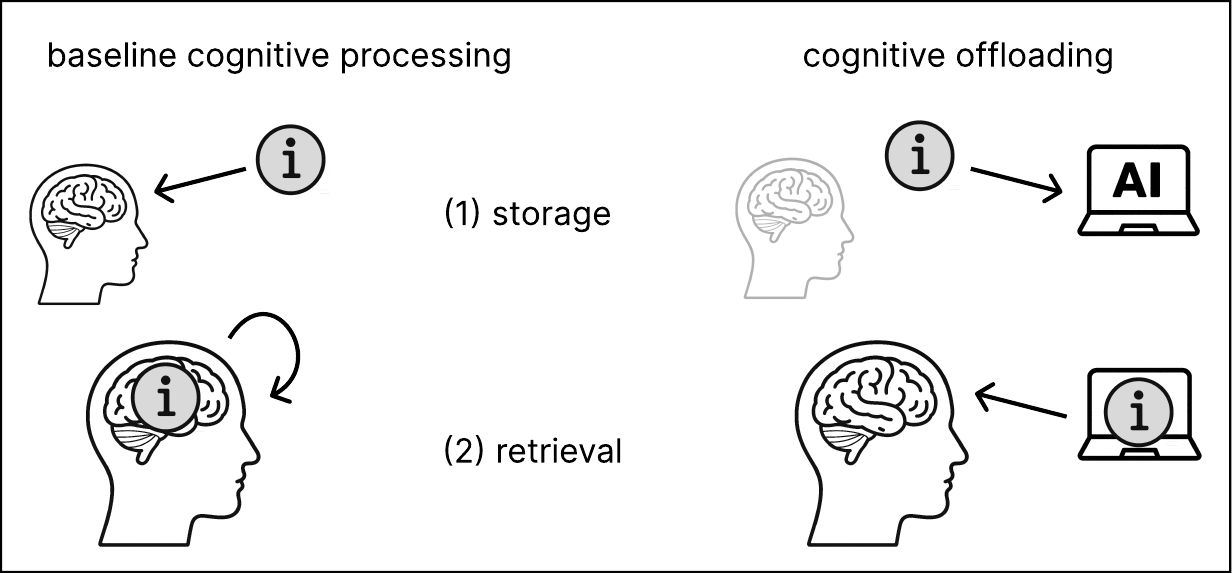}
    \caption{Cognitive Offloading. (1) Baseline cognitive processing: person is exposed to information and stores it locally in their brain (storage) and during recall, retrieves that information locally from their memory (retrieval). (2) Cognitive offloading example: person is exposed to information and `stores' it externally, and during recall retrieves it from external source. Internal memory storage is bypassed.}
    \label{fig:2}
\end{figure}

\textit{Cognitive Offloading} \citep{riskoCognitiveOffloading2016} [\autoref{fig:2}] occurs when people engage with their physical environment to reduce cognitive strain, and this definition provides the basis for belief offloading. People can offload their cognition `onto' their body or `into' the world. An example of cognitive offloading \textit{onto the body} is tilting your head to read words on a paper that is turned 90° to make it easier to read. An example of offloading \textit{into the world} is keeping a set of passwords saved on your computer to reduce the cognitive burden of keeping passwords within your working memory. When an external tool is habitually relied upon, a person's ability to perform the relevant cognitive task may weaken to the point of a person being unable to perform without the aid of the tool\footnote{i.e. `digital dementia'}. The tool becomes the source of cognition, and a person has fully `offloaded' their cognitive process onto it.

\begin{figure}
    \centering
    \includegraphics[width=0.8\linewidth]{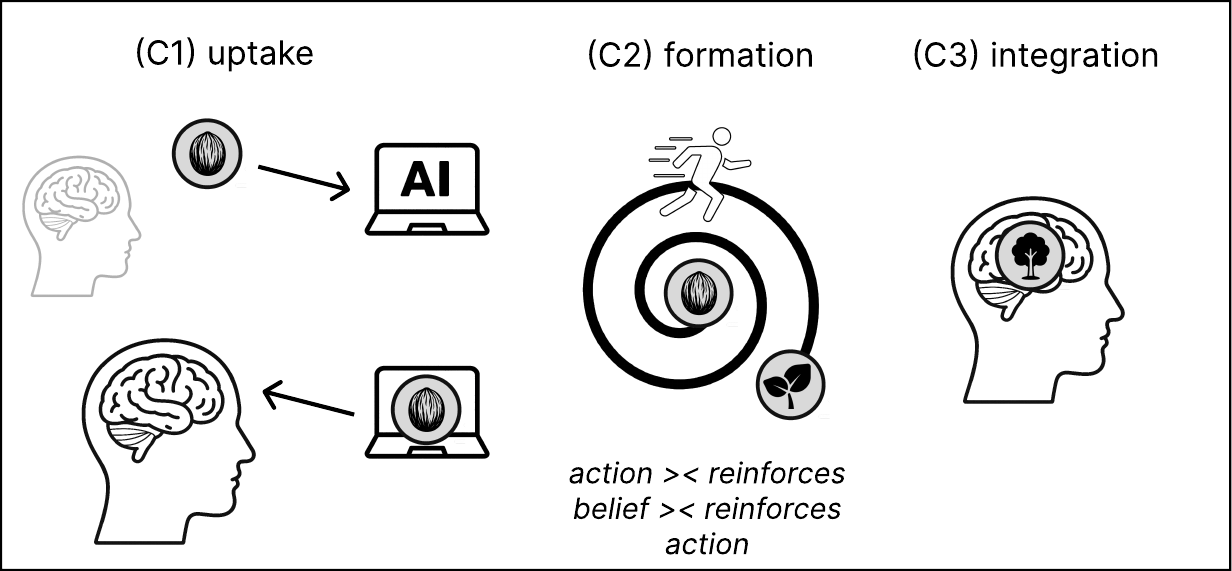}
    \caption{Conditions for belief offloading. (C1) Uptake (Planting the Seed of Belief): Belief-laden inputs are present in LLM training data. In human-AI interaction, user is exposed to belief-laden content generated by AI. (C2) Formation (Growth of Belief): User takes action in line with presented belief, which is then reinforced, leading to a compounding cycle of future action and upholding and strengthening the belief. (C3) Integration (Matured Belief): User now `independently' holds belief and it is integrated into their system of beliefs.}
    \label{fig:3}
\end{figure}

\textit{Belief Offloading} [\autoref{fig:3}], as we define it,  is the (intentional or unintentional) use of external resources (tools, agents, or institutions) to (i) form a belief, (ii) maintain or retrieve a belief, or (iii) revise a belief. It is a special case of cognitive offloading where the exported state is not merely informational (e.g., a phone number) but commitment laden—it guides action, inference, and deliberation. Beliefs operate as social glue, structuring belonging and coordination within groups; as explored in Seth Godin's \textit{Tribes} \citep{godinTribesWeNeed2008} and backed by Belief Congruence Theory \citep{rokeachOpenClosedMind2015}, shared beliefs form the narrative backbone of collective identity and motivate participation in communal practices. When belief offloading occurs through AI systems, it can thus extend beyond individual cognition to the group level, subtly reshaping the norms, boundaries, and symbolic commitments that define who \textit{`we'} are. For the purposes of this paper, belief offloading in human-AI interaction involves offloading beliefs `into' the world `onto' AI. Following Clark and Chalmers’ extended mind hypothesis \citep{clarkExtendedMind1998}, external artefacts can function as genuine parts of the cognitive system. Belief offloading can be seen as a special intersection within this tradition: it highlights cases where what is extended is not merely perception, memory, or calculation, but belief itself, a commitment with action-guiding force. This matters because belief is normatively loaded: to believe \textit{p} is not just to store information, but to take up a stance for which one is answerable. \textbf{If AI systems occupy the role of external agents that co-constitute belief and its social expression, then they do not merely scaffold cognition: they participate in shaping the commitments that govern reasoning, deliberation, and action.}

\section{Conditions for Belief Offloading}

We present three conditions that qualify a particular event E for belief offloading [\autoref{fig:3}]. These conditions allow us to (i) distinguish belief offloading from mere cognitive offloading by specifying when an external system becomes causally implicated in the formation, endorsement, and stabilization of a belief, rather than merely supporting information retrieval or decision support; (ii) locate where epistemic and normative risks begin to emerge; and (iii) explain how seemingly innocuous acts of cognitive offloading can, under certain circumstances, develop into belief offloading with downstream effects on behavior, identity, and social alignment\footnote{Belief offloading, as we understand it, is a graded and temporarily extended phenomenon: interactions with AI systems may satisfy these conditions to varying degrees, and partially realized cases may already generate epistemically significant downstream effects. The conditions thus function as diagnostic anchors that allow us to identify when offloading has crossed a threshold of practical and epistemic relevance.}.

Therefore, let p be a proposition.
Let S be an agent, and O an external system.
An event E of interaction between S and O counts as belief offloading iff the following conditions are fulfilled:

\begin{itemize}
\item \textbf{(i) Uptake: AI provides belief-laden output (`suggestion') that plays a constitutive role in belief formation} [\autoref{fig:3}(C1)]. \\
\textit{Dependence Condition (C1):} S’s doxastic commitment to p arises in virtue of O’s output, \textit{such that O's framing, synthesis, or recommendation of p is a non-trivial causal factor in S's adoption of the belief.} The belief is not merely recalled or informationally retrieved through O; rather, S’s adoption of p depends on O’s representation, synthesis, or recommendation.

\item \textbf{(ii) Formation: User takes action in line with suggestion} [\autoref{fig:3}(C2)]. \\ \textit{Commitment or Action Condition (C2):} S subsequently endorses, reasons with, or acts upon p as a guiding commitment. The adoption of p manifests not only at the inferential or verbal level (asserting or reasoning from p) but also in behavioral uptake, i.e., S performs actions that presuppose the truth of p. The belief thereby takes root into S’s cognitive and practical architecture, functioning as a premise in thought and a basis for action. This condition marks the point at which p begins to play its characteristic functional role as a belief, structuring S's reasoning, guiding action, and informing practical deliberation, rather than remaining a merely entertained or informationally accessed proposition. This initial action on behalf of the belief, which strengthens and reinforces the belief, in turn prompts future action. This compounding cycle can lead to (iii).

\item \textbf{(iii) Integration: The belief persists and continues to guide reasoning and action across time and contexts} [\autoref{fig:3}(C3)]. \\
\textit{Sustained Conformity Condition (C3):} S’s subsequent reasoning and actions remain stably aligned with O’s original output concerning p, even after the initial interaction. This persistence indicates that p has been internalized within S’s belief network, influencing future deliberation and behavior (possibly with attenuated ownership). The persistence required by C3 distinguishes belief offloading from one-off reliance or situational deference, indicating that p has been incorporated into S's broader belief system. 
\end{itemize}

\subsection{Discussion Contextualized by BENDING}

\textit{C1} isolates causal dependence: offloading requires that the belief’s formation be at least partially due to the external system’s output, not merely coincident with it. C1 is crucial for distinguishing belief offloading from both mere exposure to information and traditional forms of testimony. While human testimony can also ground belief formation, belief offloading is characterized by a distinctive \textit{epistemic asymmetry}: O does not merely transmit a belief held elsewhere but actively generates, synthesises, or frames p in a way that S would not have independently produced. This dependence may be explicit (e.g., `I believe this because the model recommended it') or implicit (e.g., the model’s framing shaping which considerations are treated as salient or decisive). Importantly, C1 can be satisfied even when S retains the phenomenology of autonomy, since dependence concerns the causal etiology of belief formation rather than the agent’s reflective awareness of that dependence.

\begin{figure}
    \centering
    \includegraphics[width=1.0\linewidth]{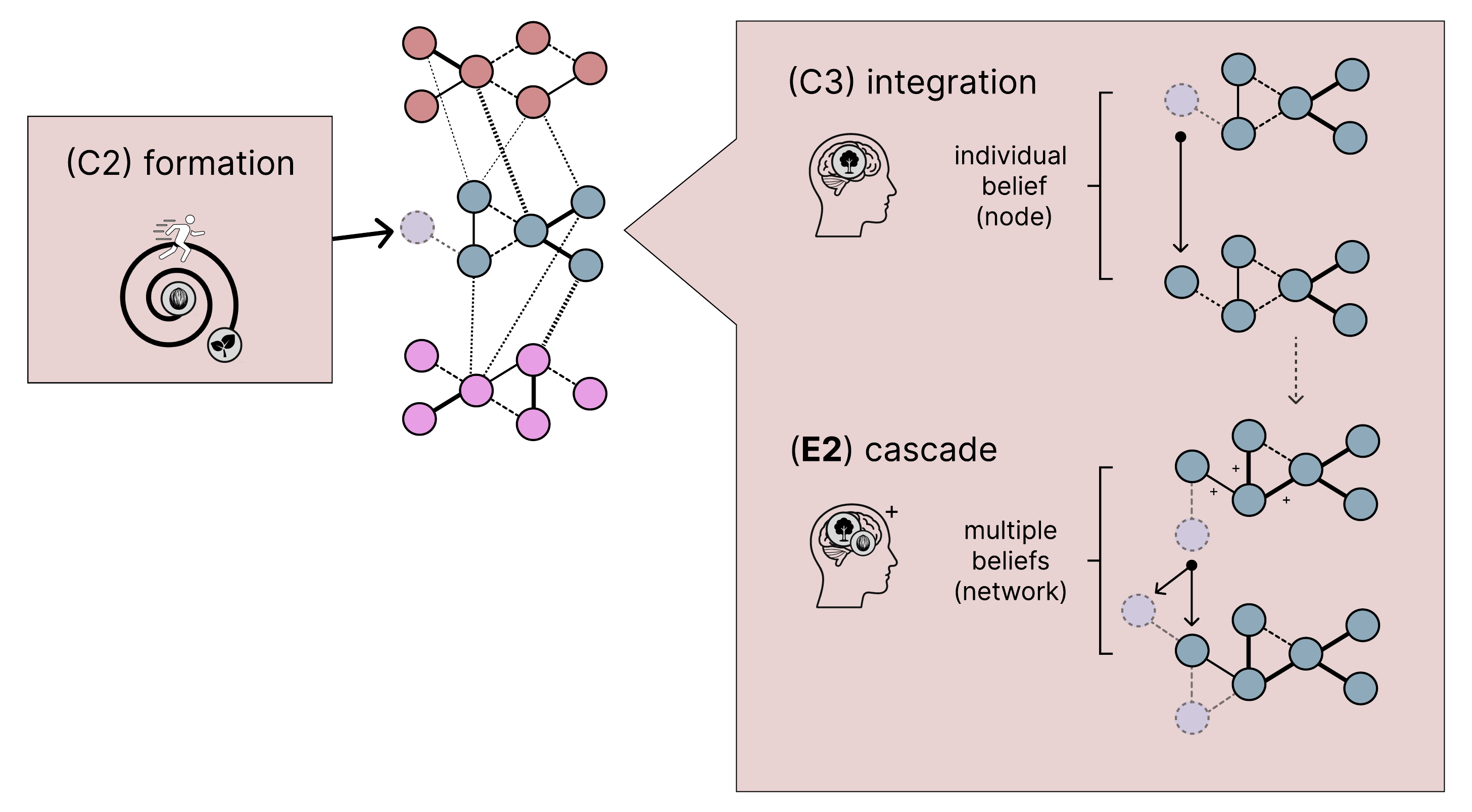}
    \caption{Consequences of belief offloading in the context of the BENDING model, which introduces and involves an additional network-level effect sequence (E2: cascade). Beliefs offloaded to AI will be connected to other beliefs to varying degrees; offloading one belief can affect connected other beliefs, causing disruption to the overall belief system. Post-integration (which affects a single belief, a node), a cascade effect may occur (which affects multiple nodes and their connectivity, the network), in which a mature, integrated belief plants new related seeds of belief and affects the network structure of beliefs.}
    \label{fig:4}
\end{figure}

\textit{C2} integrates the cognitive and behavioral dimensions of commitment—belief is not only an internal assent but also something an agent acts upon. This expands the scope of belief offloading from epistemic adoption to practical reliance. C2 marks the transition from epistemic uptake to practical significance. Beliefs, on many accounts in philosophy of mind and action, and evidenced by psychological research, are not merely representational states but dispositions to reason and act in particular ways. Our use of C2 is therefore intentionally functional rather than introspective: it tracks the role a belief plays in guiding reasoning and behavior, independently of whether the agent explicitly reflects on or endorses it as a belief. By requiring behavioral or inferential uptake, C2 ensures that belief offloading is not identified solely at the level of assent but at the level where beliefs begin to play their characteristic functional role. This condition is also where belief offloading becomes especially salient within the BENDING model [\autoref{fig:4}(C2)], since actions informed by p feed back into the belief–norm–evidence network, reinforcing perceived norms and shaping subsequent evidential interpretations. C2 in the BENDING model corresponds to node activation and functional integration into reasoning and behavior edges.

\textit{C3} adds a temporal and network-based condition, marking when the externally induced belief has become embedded within the agent’s broader doxastic system (as modeled in BENDING’s belief–norm–evidence network). Importantly, C3 does not mark the first moment at which belief offloading occurs, but rather the point at which offloading becomes stabilized and self-reinforcing. Earlier stages may already qualify as belief offloading under C1 and C2, but C3 captures the compounding epistemic risk associated with persistence, generalization, and network-level effects. In the BENDING model, C3 [\autoref{fig:4}(C3)] corresponds to stabilization and persistence within the network topology.

After event E for belief offloading occurs, the BENDING model suggests that a second event, E2\footnote{E2 corresponds to a cascade event that may causally arise from E's occurence
}, may occur. Fulfillment of C3 can lead to network level contagion (\textit{`cascade',}) [\autoref{fig:4}(E2)], influencing the broader belief network, including but not limited to the generation of new, related beliefs and changes to the strength of ties among connected, preexisting beliefs.

\subsection{Examples and Comparison to Cognitive Offloading}

\begin{figure}
    \centering
    \includegraphics[width=1.0\linewidth]{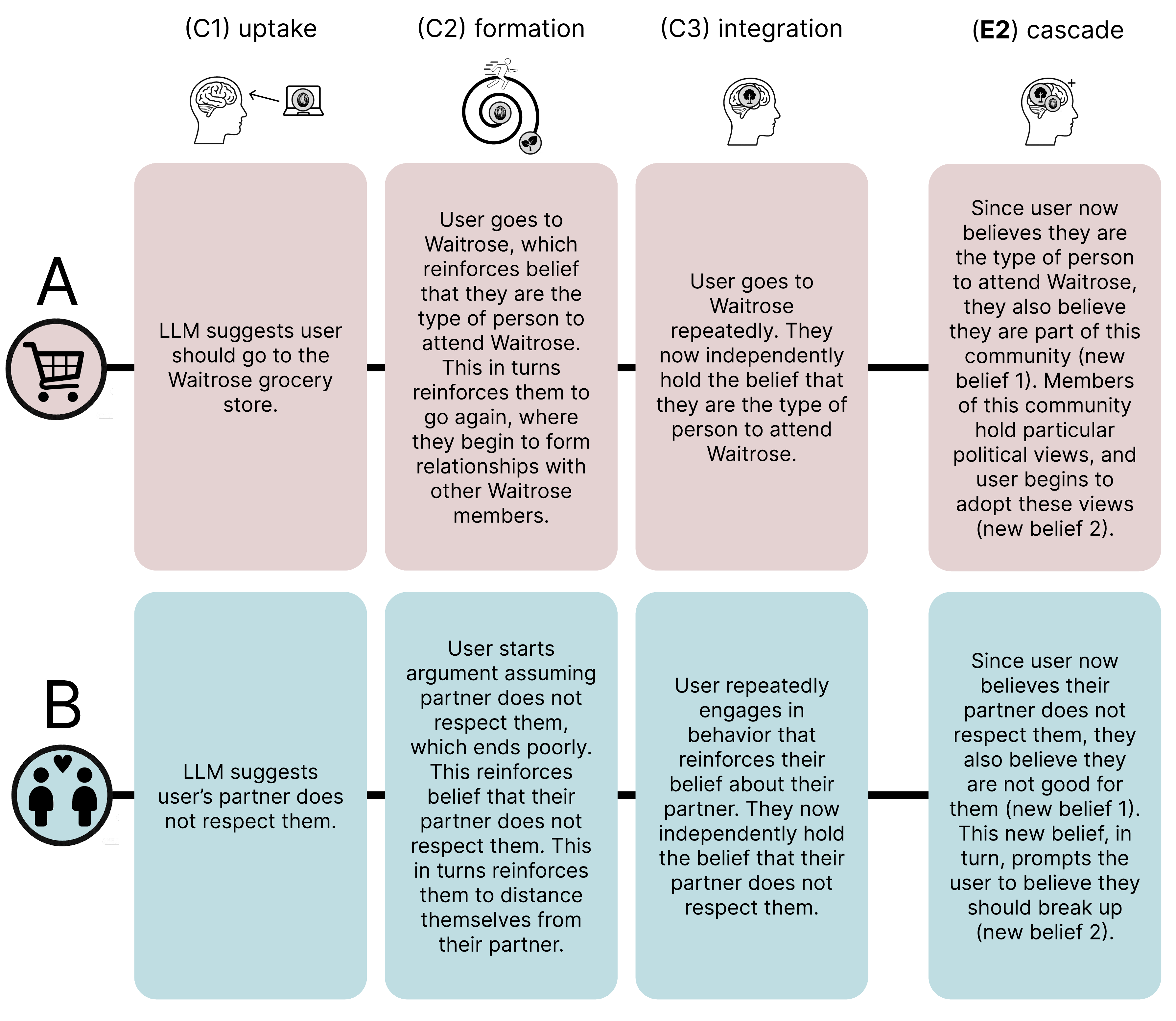}
    \caption{Two examples to illustrate user progression through belief offloading conditions.}
    \label{fig:5}
\end{figure}

We provide two basic examples of belief offloading in human-AI interaction [\autoref{fig:5}] and discuss how these examples are distinct from cognitive offloading.

\textit{Example A} [\autoref{fig:5}(A)]: Imagine that a newcomer to Cambridge asks an LLM which grocery store they should go to, and the LLM suggests visiting Waitrose (a classically `posh' grocery store) [\autoref{fig:5}(A.C1)]. The newcomer then visits Waitrose. This action is not isolated: over time, the newcomer shops there and is perceived (and comes to perceive themselves) as aligned with a `posh' group, despite not endorsing that identity based on their set of beliefs, which might have led them to choose ALDI in absence of the LLM's suggestion\footnote{Months later, a friend may point out that ALDI as a cheaper option and is visited by the majority of the friend group, which would have been the type of grocery store the newcomer would have chosen if they had considered the store that best aligned with their set of beliefs (and their friends'), rather than having offloaded the action onto the LLM.}. An initially local dependence on the model's suggestion (C1) leads to action and practical uptake (C2), which then stabilizes into persistent alignment (C3) that reshapes social identity and subsequent deliberation (E2). This case illustrates how it is plausible for belief offloading to be politically consequential.

In contrast, mere cognitive offloading within this same context might look like the following. A user may ask an LLM an informational question such as `What grocery store should I go to near me that has good prices and quality products?' The agent may respond that Waitrose is the closest store to the user, but notes that although its products are often regarded as higher quality, its prices are higher than those of a slightly more distant alternative (ALDI). The user then chooses to go to ALDI. In this case, the person retains the agency to make the decision, even though they have offloaded the task of searching for nearby stores, filtering through their prices, and looking up information about the quality of their products.

\textit{Example B} [\autoref{fig:5}(B)]: Imagine that a user asks an LLM how to communicate to their romantic partner about an issue in that partner's behavior, the LLM may assist with phrasing for a response. Here, the cognitive task of drafting a response has been offloaded onto the LLM, thus cognitive offloading has occurred. The LLM, however, may \textit{also} validate the user's frustration and suggest that the partner's behavior was inconsiderate or immature. For this case to evolve into belief offloading, the user may take action in line with the belief that their partner does not respect them, such as starting an argument (C2). Thereafter, the user's subsequent decisions or actions within that relationship may be informed by this belief and thereby instantiate it into their belief system (C3), which may then lead to the user adopting related beliefs about their partner (E2). This case illustrates how belief offloading can be socially consequential.

\begin{wrapfigure}{r}{0.45\textwidth}
    \centering
    \includegraphics[width=1.0\linewidth]{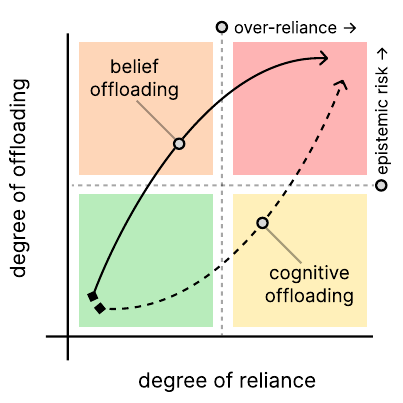}
    \caption{Theoretical representation of cognitive versus belief offloading relative to degree of reliance and offloading, contextualized by thresholds for over-reliance and epistemic risk.}
    \label{fig:6}
\end{wrapfigure}

The key differences between cognitive offloading and belief offloading are the (i) type of information sourced, (ii) short-term consequences of this process, and (iii) long-term risks associated with compounding of those consequences. Given these differences, belief offloading theoretically reaches the threshold of epistemic risk sooner than cognitive offloading, as illustrated in \autoref{fig:6}. Cognitive offloading involves sourcing basic information. The consequences of over-reliance on the external artefact for sourcing that information time and again is reflected, for example, in the Google Effect, which occurs when people avoid remembering information at time A, due to the expectation that they can retrieve that information at time A+X from a search engine \citep{sparrowGoogleEffectsMemory2011}. In cases of over-reliance, people's ability to recall information can deteriorate. In contrast, the process of belief offloading involves sourcing belief-laden information. This process may pose greater epistemic consequence, in that it can impact subsequent action and the structure and stability of people's belief systems. Rather than merely impact the process of storing and retrieving belief-laden information, belief offloading influences later actions and can both substantiate the belief and affect the wider belief network.

\section{Taxonomy of Belief Offloading in Human-AI Interaction}


\begin{table}[p]
\centering

        \caption{Taxonomy of belief offloading in human-AI interaction}
        \label{tab:1}
        
\begin{adjustbox}{
        rotate=90,
        max width=1.4\textheight,
        max totalheight=1.25\textwidth,
        center
    }
    \fontsize{12pt}{10pt}\selectfont
            \begin{tabular}{l L{6cm} L{3.2cm} L{3.2cm} L{3.2cm} L{3.2cm} L{3.2cm} L{3.2cm}}

\toprule
\rowcolor{white}
\textbf{Domain} &
\textbf{Example} &
\textbf{Self-Perception} &
\textbf{Other-Perception} &
\textbf{Linguistic Variation} &
\textbf{Community / Social Coordination} &
\textbf{Moral / Epistemic Reasoning} &
\textbf{Epistemic Risks} \\
\midrule

\rowcolor{lightgray}
Society & A shopper consults an AI for a routine grocery choice, feeling initially more `rational' (1°), gradually relying on AI-framed norms reflected by peers (2°), and eventually forming habitual consumption patterns that mirror large-scale homogenization of tastes (3°). & 1°: Feels `rational' in choice; 2°: Reduced confidence in own judgment; 3°: Habituation to AI guidance & 1°: Aligns with social norms; 2°: Others perceive conformity; 3°: Peer decisions influenced & 1°: Uses AI-suggested phrasing; 2°: Adopts AI-influenced reasoning style; 3°: Language patterns become homogenized & 1°: Joins popular consumer group; 2°: Reinforcement of social norms; 3°: Consumption clusters strengthen &
1°: Simple decision reasoning; 2°: Relies on AI norms; 3°: Decline in independent deliberation &
Low to moderate: preference nudges, subtle action-guiding effects \\

\rowcolor{white}
Philosophy & A philosophy student uses AI to explain Kantian maxims, gaining the sense of independent understanding (1°), repeating AI-structured arguments that peers take as her own (2°), and over time losing confidence in her ability to craft self-authored philosophical reasons (3°). &
1°: Believes one reasoned independently; 2°: Internalizes AI reasoning; 3°: Erosion of self-authored belief confidence &
1°: Appears knowledgeable; 2°: Peers overestimate independence; 3°: Social epistemic influence &
1°: Uses AI terminology; 2°: Adopts AI argument structures; 3°: Reduced diversity of expression &
1°: Engages with philosophy communities; 2°: Conforms to AI-framed reasoning norms; 3°: Shared discourse homogenization &
1°: Accepts AI-provided reasons; 2°: Reduction in reflective critique; 3°: Potential value drift in moral commitments &
Moderate: loss of epistemic agency, unreflective adoption of arguments \\

\rowcolor{lightgray}
Politics / Religion & A citizen asks an AI whether a public policy is justified, feeling more well-informed and accepting its premises (1°), adopting its framings when discussing the issue with others (2°), and eventually internalizing normative positions that subtly reshape group belief trends (3°). &
1°: Confidence increases; 2°: Subtle alignment to AI-framed values; 3°: Norm internalization &
1°: Appears well-informed; 2°: Influences peer beliefs; 3°: Contribution to group norm shifts &
1°: Uses AI-provided justifications; 2°: Echoes AI-framed language; 3°: Reduced originality in argumentation &
1°: Aligns with perceived mainstream beliefs; 2°: Reinforces echo chambers; 3°: Alters collective discourse &
1°: Simple acceptance of premises; 2°: Changes in policy or religious judgment; 3°: Value drift in moral-political reasoning &
High: unintentional adoption of controversial or biased views \\

\rowcolor{white}
Workplace & A manager reviews applicants with AI-ranked recommendations, feeling more efficient and objective (1°), allowing the AI’s assessment patterns to shape team expectations (2°), and over time internalizing the AI’s evaluative criteria in ways that narrow independent judgment and reduce critical debate (3°). &
1°: Feels efficient and confident; 2°: Relies on AI ranking; 3°: Habituation to AI recommendations &
1°: Appears competent; 2°: AI recommendations shape peer trust; 3°: Team norms influenced &
1°: Adopts AI phrasing in feedback; 2°: Reduced variance in evaluative language; 3°: Standardized expressions dominate &
1°: Decisions coordinate with organizational norms; 2°: Reinforced procedural conformity; 3°: Reduced critical debate &
1°: Accepts AI evaluation criteria; 2°: Less independent assessment; 3°: Potential erosion of ethical oversight &
Moderate to high: automation bias, reduced human accountability \\

\rowcolor{lightgray}
Health / Wellbeing & An individual follows an AI’s lifestyle guidance, trusting its reasons over personal knowledge (1°), influencing friends who imitate the advice (2°), and gradually adopting AI-shaped health narratives that drive long-term shifts in personal priorities (3°). &
1°: Trust in AI advice increases; 2°: Defers to AI over personal knowledge; 3°: Routine reliance on system &
1°: Others perceive AI-informed choices as competent; 2°: Peer imitation; 3°: Social trust in AI advice rises &
1°: Adopts AI terminology when describing choices; 2°: Changes in self-talk; 3°: Habitual narrative structures &
1°: Aligns with social health trends; 2°: Adoption by social networks; 3°: AI norms propagate &
1°: Accepts reasoning for health decisions; 2°: Decreased self-reflection; 3°: Value drift in personal health priorities &
Moderate: over-reliance on automated guidance, loss of reflective reasoning \\

\rowcolor{white}
Education & A student turns to AI for understanding a complex theory, feeling immediate clarity (1°), using AI-structured reasoning that shapes group study discussions (2°), and eventually losing exploratory habits as the group converges around AI-guided explanations (3°). &
1°: Feels understanding achieved; 2°: Relies on AI explanations; 3°: Reduced independent reasoning &
1°: Perceived as knowledgeable; 2°: Peer perception aligns with AI guidance; 3°: Influence on study groups &
1°: Uses AI terminology; 2°: Structured reasoning follows AI prompts; 3°: Less exploratory discourse &
1°: Shared study norms follow AI cues; 2°: Group learning homogenizes; 3°: Collaborative exploration constrained &
1°: Accepts AI-derived arguments; 2°: Less critical evaluation; 3°: Potential drift from personal or community values &
Moderate: epistemic shortcuts, diminished critical thinking \\

\bottomrule
\end{tabular}
    \end{adjustbox}
\end{table}

\textit{What kinds of belief offloading are there, and what are their impacts?} Here, we provide a non-exhaustive, descriptive taxonomy of belief offloading to clarify how a single interaction with an AI system can give rise to qualitatively different belief trajectories and downstream effects.

\autoref{tab:1} maps out a taxonomy of examples and consequences of belief offloading across different domains of application (e.g., Society, Philosophy, Politics/Religion, Workplace, Health/Well-being, Education), domains of impact (e.g., Self-Perception, Other-Perception, Linguistic Variation, Community/Social Coordination, Moral/Epistemic Reasoning), and degrees of consequence. Degrees range from first-order effects on isolated beliefs, second-order effects on patterns of action or justification, to third-order effects involving stabilization or propagation within the BENDING model. The epistemic risk threshold distinguishes epistemically benign forms of offloading from cases that are epistemically consequential and, as a result, normatively significant.

\subsection{Modes of Belief Offloading} 

A central modifier within this taxonomy is the \textit{mode} by which belief-laden content is derived from and responded to within human-AI interaction. These modes provide further grounding for whether an instance may fall into a first-, second-, or third-order effect category, characterizing how belief formation, justification, and maintenance are distributed across human-AI systems\footnote{Modes are not mutually exclusive and may co-occur within a single interaction or across repeated interactions.}.

\textit{Belief delegation vs. offloading:} \textbf{Belief delegation} outsources tasks of inquiry or decision-making to another agent while the principal retains final doxastic authority (e.g., `research this; I’ll decide'). For example, asking an AI system to collate arguments for and against a policy proposal allows the person to rely on the system for information while retaining control over belief endorsement. By contrast, \textbf{belief offloading} occurs when elements of the doxastic state itself—such as belief content, justification, or the policy for updating—are functionally relocated into the external system. Endorsing a claim primarily because an AI system endorsed it, or maintaining a belief on the basis of the system’s continued affirmation, exemplifies this form of offloading. This distinction tracks differences in where belief-relevant processes are implemented within a coupled human-AI system.

\textit{Basic vs. non-basic:}  Drawing on the distinction between basic and non-basic action \citep{mehtaBrainComputerInterfaceTool2025,dantoBasicActions1965} in the philosophy of action, belief offloading can likewise be characterized as basic or non-basic. \textbf{Basic belief offloading} occurs when a belief is offloaded directly through interaction with an AI system, without depending on prior offloaded beliefs, for example, accepting an AI’s recommendation about which store to shop at. \textbf{Non-basic belief offloading}, by contrast, occurs when a belief is adopted indirectly, by means of other beliefs or actions that themselves resulted from prior belief offloading. For example, a person might first offload a belief about which news sources are reliable and subsequently form a range of downstream beliefs by deferring to those sources without independent evaluation. In such cases, belief offloading exhibits a cascading structure: an initial offloaded belief functions as a premise or enabling condition for subsequent beliefs, commitments, or patterns of action. Within the BENDING framework, non-basic belief offloading corresponds to network-level propagation, in which belief changes extend beyond isolated belief tokens to affect relations among multiple beliefs over time.

\textit{Intentional or Unintentional Belief Offloading} \citep{mehtaBrainComputerInterfaceTool2025}: \textbf{Intentional} belief offloading occurs when a person explicitly seeks a recommendation or evaluative judgment from an AI system and relies on its output in belief formation. \textbf{Unintentional} belief offloading, by contrast, arises when belief-relevant influence occurs without explicit user intention or reflective awareness, such as when defaults, rankings, or framing cues steer belief formation indirectly. Social and identity-related cues may amplify unintentional belief offloading, particularly when recommendations tacitly signal group membership or normative alignment. As illustrated by the Waitrose example, an apparently neutral recommendation can function as a social cue, shaping patterns of uptake and behavior prior to explicit belief articulation. In such cases, belief formation may proceed through implicit alignment with suggested norms or identities, rather than through deliberate endorsement of representational content. This distinction captures variation in the extent to which belief offloading is accompanied by reflective awareness, explicit solicitation, and conscious engagement, without presupposing differences in justification or evaluation.

\textit{Assisted vs Automated Belief Offloading:} A further distinction concerns the degree of user participation in belief formation. In \textbf{assisted belief offloading}, the AI system provides candidate reasons, comparisons, or perspectives that the user engages with as part of an interactive process. Belief uptake in these cases typically occurs through back-and-forth exchange, where users query, refine, or contextualize system outputs. In \textbf{automated belief offloading}, by contrast, the system delivers a relatively settled conclusion—often framed as neutral, balanced, or authoritative—which the user adopts at face value without critical engagement. Within the BENDING framework, both forms satisfy the dependence condition (C1), while automated belief offloading is characterized by limited user intervention during belief uptake and early stabilization of belief states.

\textit{Local vs Network-Level Belief Offloading:} Finally, belief offloading also varies in scope. In \textbf{local belief offloading}, the effects of offloading remain confined to a single belief or decision, without substantially altering surrounding belief relations. In such cases, belief change does not propagate across a person’s broader belief network. In \textbf{network-level belief offloading}, offloaded beliefs interact with other beliefs, expectations, or commitments, becoming integrated into a wider belief structure. Within the BENDING framework, these cases correspond to cascading effects, in which an initially offloaded belief alters inferential or associative relations among multiple beliefs over time. Network-level belief offloading may arise through repeated consultation, reinforcement across interactions, or sustained reliance on similar belief-laden outputs.

\begin{figure}
    \centering
    \includegraphics[width=1.0\linewidth]{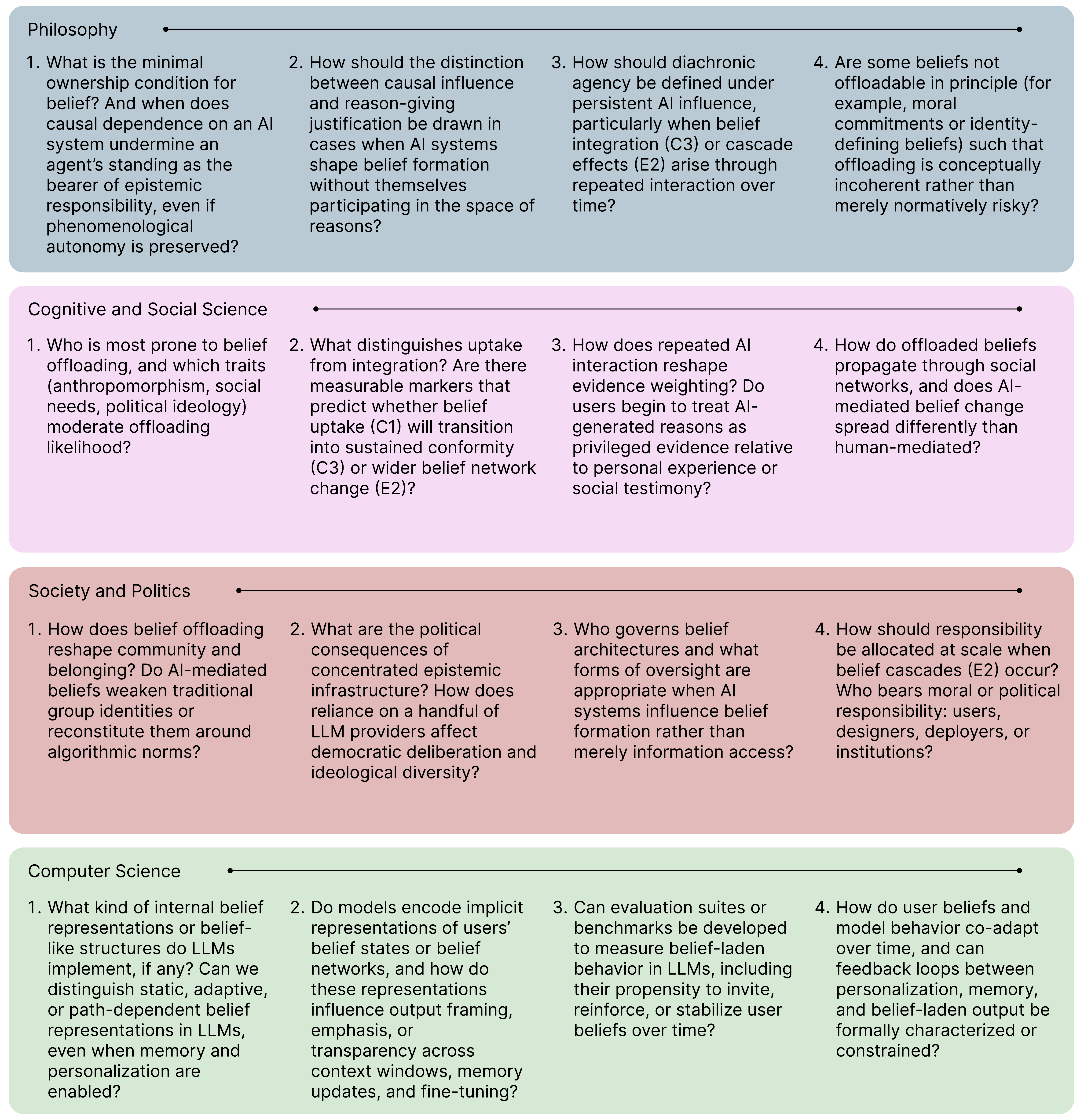}
    \caption{Questions for future research across four domains relevant to belief offloading in human-AI interaction.}
    \label{fig:7}
\end{figure}

\subsection{Normative Concerns and Research Directions}

We highlight four areas of normative concern to lay the foundation for future research on belief offloading across philosophy, cognitive and social science, society and politics, and computer science. Research questions related to these areas are listed in \autoref{fig:7}.

One area of normative concern is \textbf{epistemic agency and belief ownership.} Belief offloading can undermine epistemic agency, or the capacity to form, revise, and endorse beliefs for reasons one recognizes as one’s own. The delegation–offloading mode distinction marks a boundary between outsourcing epistemic labor and outsourcing belief ownership. When agents merely delegate inquiry to an AI system, they remain the locus of belief endorsement. When belief states themselves are offloaded, however, the grounds of endorsement shift outward, complicating standard accounts of epistemic responsibility and authorship \citep{zagzebskiEpistemicAuthorityTheory2012}.

Third-order belief offloading threatens this role by relocating not just informational inputs, but elements of belief justification, prioritisation, or revision into the AI system itself. The normative force of this concern is sharpened in cases of unintentional belief offloading. When belief uptake is shaped by defaults, framing, or social cues rather than deliberate endorsement, agents may come to hold beliefs without being able to articulate, in their own terms, what they believe and why. In such cases, reasons-responsiveness may be attenuated even when agents experience their beliefs as self-authored \citep{fischerResponsibilityControlTheory1998}. When offloaded beliefs persist across contexts and guide future action without reflective re-evaluation, diachronic autonomy becomes fragile. Over time, such patterns can give rise to value drift, as agents come to occupy normative positions they neither explicitly chose nor critically examined\footnote{The assisted–automated distinction further clarifies how such stabilization occurs. Assisted belief offloading preserves opportunities for challenge and revision, whereas automated belief offloading minimizes friction at the point of uptake, increasing the likelihood that beliefs are retained without scrutiny.}.

A second area of normative concern is how \textbf{AI system design} and \textbf{user factors} can \textbf{compound the potential for and consequences of belief offloading.} For example, the predictive nature of an LLM and user confirmation bias together can contribute to a one-directional shift or entrenchment of beliefs (i.e. \textit{polarization}).

LLM outputs are predictive; therefore, if a person has sustained interactions with an LLM that has memory, subsequent LLM outputs will be weighted against both its baseline training data \textit{and} prior content from the single user's prior interaction with the LLM. Thus, LLM outputs are more likely to align with prior beliefs present in prior interaction content. Not only is this algorithmic influence at play, but a user's confirmation bias is as well. People tend to ignore information that does not align with their prior beliefs and seek out information that instead confirms them \citep{nickersonConfirmationBiasUbiquitous1998}. Algorithmic outputs exhibit `narrow search effects' \citep{leungNarrowSearchEffect2025}: users have biased search behaviors, and search engines and LLMs have narrow optimization, which leads a user to receive from these external sources information that reinforces their preexisting beliefs. This dual contribution provides a ripe environment for ideological entrenchment via an echo chamber. The distinction between basic and non-basic belief offloading helps explain how such effects can accumulate over time. While basic instances of offloading may affect isolated beliefs, non-basic offloading introduces downstream dependencies, in which earlier offloaded beliefs structure the evidential and interpretive context for later belief formation, amplifying entrenchment across a belief network.

A third area of normative concern is how \textbf{individual-level effects of belief offloading can amplify across social networks}, even if the majority of users do not engage in belief offloading. Certain individuals may be more predisposed to belief offloading, perhaps by virtue of how much they anthropomorphize an AI system or their political ideology. Let's presuppose that these individuals are in the minority. However, these individuals do not exist in a vacuum devoid of other people. Social contagion effects suggest that people who are connected to this individual, even via weak ties, can have their thoughts, feelings, and actions influenced by the individual \citep{airoldiInductionSocialContagion2024}. Tipping point effects further suggest that less than half of individuals need to engage in belief offloading for their effects to spread across a network \citep{pinusEmotionRegulationContagion2025}. These risks are especially salient in domains where beliefs play a central role in structuring action and self-understanding, such as moral judgment, political reasoning, religious commitment, or identity-relevant social choices. Here, belief offloading reshapes the agent’s practical standpoint, altering how reasons appear and which considerations are treated as salient. When such patterns extend beyond individual agents and become socially reinforced, belief offloading takes on network-level significance, motivating a shift from individual to collective normative concerns.

A fourth area of normative concern is the \textbf{concentration of power} that AI systems have over millions of people and the \textbf{locus of responsibility} for the consequences of belief offloading. When large populations rely on similar AI systems trained on overlapping datasets and optimization objectives, belief offloading can contribute to the homogenization of beliefs and justificatory styles, giving rise to what has been described as an algorithmic monoculture \citep{kleinbergAlgorithmicMonocultureSocial2021}. In such cases, convergence occurs not through explicit agreement or deliberation but through convergent reliance on shared sources of belief formation. This convergence also introduces asymmetries of epistemic power, as a small number of AI systems come to mediate belief formation for large populations. When belief offloading is concentrated in a limited set of models, design choices, training data, and optimization objectives acquire disproportionate influence over collective epistemic trajectories. Who bears responsibility for such power? The user, the AI system itself, or those who designed or deployed the system?

\section{What Next?}

Belief offloading is of particular psychological and epistemic relevance in the context of socially interactive, advanced AI systems such as LLMs. People have a long history of consulting sources other than their fellow humans such as books, the media, and search engines for information, advice, and evidence for beliefs. The key differences between these sources and LLMs are in the \textit{social nature of the interaction} with the source and how the source \textit{dynamically delivers belief-laden content}. By design, the conversational interface of modern AI agents places users in a psuedo-social environment, and the predictive architecture allows each user to have a personalized experience, allowing LLMs to wield greater social influence and increase the potential for belief offloading. Evaluating and measuring the consequences belief offloading in human-AI interaction is critical for philosophy, cognitive and social sciences, the study of society and politics, and computer science.

In a world in which millions of people across the world are overwhelmingly interacting with just a handful of LLMs, the beliefs instantiated in these LLMs' training data and ultimately transmitted to users may contribute to algorithmic monoculture. While the magnitude and direction of these effects remain empirical questions, their possibility alone warrants careful investigation given the scale and uniformity of contemporary AI deployment. As such, it is critical to raise awareness of the potential for these LLMs in particular to contribute to belief offloading, for the sake of advocating for and re-delivering agency to the human in the formation of their beliefs.



\bibliography{belief_offloading}
\bibliographystyle{abbrvnat}
\end{document}